\documentclass[twocolumn]{article}
\usepackage[utf8]{inputenc}
\usepackage{geometry}
\usepackage{booktabs}
\usepackage{tabularx}
\usepackage{subcaption}
\usepackage{float}
\usepackage{hyperref}
\usepackage{multirow}
\usepackage{makecell}
\usepackage{graphicx}
\usepackage[numbers]{natbib}
\usepackage{xcolor}
\usepackage{stfloats}
\usepackage{authblk} 

\title{Improving Low-Resource Translation with Dictionary-Guided Fine-Tuning and RL: A Spanish-to-Wayuunaiki Study}

\author[1]{Manuel Mosquera}
\author[1]{Melissa Robles}
\author[1]{Johan Rodríguez}
\author[1]{Rubén Manrique}
\affil[1]{Universidad de los Andes, Bogotá, Colombia\\
\texttt{\{ma.mosquerao, mv.robles, jd.rodriguez1234, rf.manrique\}@uniandes.edu.co}}

\date{} 

\begin{document}

\twocolumn[
\maketitle
\begin{abstract}
Low-resource machine translation remains a significant challenge for large language models (LLMs), which often lack exposure to these languages during pretraining and have limited parallel data for fine-tuning. We propose a novel approach that enhances translation for low-resource languages by integrating an external dictionary tool and training models end-to-end using reinforcement learning, in addition to supervised fine-tuning. Focusing on the Spanish–Wayuunaiki language pair, we frame translation as a tool-augmented decision-making problem in which the model can selectively consult a bilingual dictionary during generation. Our method combines supervised instruction tuning with Guided Reward Policy Optimization (GRPO), enabling the model to learn both when and how to use the tool effectively. BLEU similarity scores are used as rewards to guide this learning process. Preliminary results show that our tool-augmented models achieve up to +3.37 BLEU improvement over previous work, and a 18\% relative gain compared to a supervised baseline without dictionary access, on the Spanish–Wayuunaiki test set from the AmericasNLP 2025 Shared Task. We also conduct ablation studies to assess the effects of model architecture and training strategy, comparing Qwen2.5-0.5B-Instruct with other models such as LLaMA and a prior NLLB-based system. These findings highlight the promise of combining LLMs with external tools and the role of reinforcement learning in improving translation quality in low-resource language settings.
\end{abstract}
\vspace{1em}
]


\section{Introduction}\label{introduction}
Natural language processing (NLP) has witnessed remarkable progress in recent years, yet such advances have largely bypassed low-resource languages, especially Indigenous languages, due to the scarcity of high-quality parallel corpora and the predominance of oral over written traditions \cite{ignat-etal-2024-solved, preserving_heritage2024}. As a result, even state-of-the-art generative AI systems struggle to produce reliable output: a BIDLab study found that AI responses in Indigenous languages are correct only 54\% of the time, with answers on average four times shorter and noticeably degraded in fluency and adequacy \cite{lucas_performance_2025}.

Against this backdrop, community-driven and academic initiatives have begun to address the gap. Notably, the AmericasNLP Shared Task (2025) introduced translation benchmarks covering 14 Indigenous languages from North, Central, and South America, catalyzing new efforts in corpus compilation, data curation, and evaluation protocols tailored for severely data-scarce contexts \citep{amercasNLP_2025}. These efforts not only facilitate digital access for largely marginalized language communities but also reinforce ongoing programs in language revitalization, educational outreach, and cultural heritage preservation.

Methodologically, most prior work on Indigenous language translation employs supervised fine-tuning of large language models (LLMs) on small, carefully curated parallel datasets \cite{amercasNLP_2025, hus-etal-2025-machine}. While such approaches have yielded promising gains in some low-resource scenarios, they remain fundamentally constrained by the availability of annotated data and tend to generalize poorly to out-of-distribution inputs \cite{ignat-etal-2024-solved, loreslm-ws-2025-1, khade-etal-2025-challenges, amercasNLP_2025}. Consequently, purely supervised paradigms struggle to capture the linguistic richness and variability inherent to Indigenous languages, which often exhibit complex morphology, dialectal variation, and limited orthographic standardization.

Recently, reinforcement learning (RL) has emerged as a promising post-training strategy, requiring far fewer annotated examples and capable of both complementing and supplanting traditional supervised techniques. These RL methods such as Proximal Policy Optimization (PPO) \cite{PPO2017} and the more recent Generalized Reinforcement Policy Optimization (GRPO) \cite{shao2024deepseekmathpushinglimitsmathematical, grpo_deepseek2025} have gained popularity in LLM training. These techniques have proven effective in aligning model outputs with human preferences, as demonstrated in Reinforcement Learning from Human Feedback (RLHF) \cite{RLFHF}, and have subsequently been employed to enhance the reasoning abilities of LLMs \cite{deepseek_r1}. In contrast to supervised fine-tuning, RL enables models to learn policies over sequences of actions, facilitating dynamic interaction with an environment and enabling better adaptation to sparse or delayed feedback.  However, RL methods have yet to be explored in the context of machine translation, particularly in low-resource settings. 

Furthermore, RL has been used to extend model capabilities through the integration of external tools that help the model with different tasks like executing code, performing math calculations, or searching the web \cite{search-r1, feng2025retool, goldie2025syntheticdatageneration}. These agent-like abilities enhance model performance in domains where specialized tools can provide meaningful support. A key advantage of RL in tool usage is that it enables models to learn autonomously how to use tools effectively to improve task performance.  Despite its effectiveness, little work has focused on developing or leveraging such tools specifically for machine translation \cite{brivaiglesias2025aiagentsnewmachine}, especially in the low-resource context.

In this paper, we propose an alternative to traditional fine-tuning strategies for improving machine translation performance in Wayuunaiki, the most widely spoken Indigenous language in Colombia. Our approach builds on the instruction-tuned model Qwen2.5-0.5B-Instruct \cite{qwen2025}, which we further train using reinforcement learning. Unlike standard methods, we frame the model as an agent capable of interacting with an external Wayuunaiki–Spanish dictionary. To support this interaction, we adopted the GRPO framework introduced by DeepSeek \cite{deepseek_r1}, enabling the model to learn when and how to call the dictionary. This agent-based formulation facilitates tool-augmented translation and reduces reliance on large annotated corpora. To the best of our knowledge, this is the first work to incorporate a dictionary as an interactive tool in low-resource machine translation, and the first to apply RL to adapt LLMs in the translation context. By framing the model as an agent, our methodology opens new avenues for research into tool-augmented translation strategies for underrepresented languages.

\subsection{Paper organization}
This paper is divided into four main sections. The Related Work section reviews existing approaches to machine translation for low-resource and Indigenous languages, emphasizing the challenges of data scarcity and highlighting recent efforts to incorporate reinforcement learning into translation. The Methods section presents our framework for tool-augmented translation, describing both the supervised fine-tuning pipeline and the reinforcement learning setup, including the GRPO algorithm, the construction of our parallel corpus, model selection, and training protocols. In the Results section, we present our experimental findings, followed by the Discussion section, which reflects on the implications of tool-augmented machine translation in low-resource settings, addresses limitations, and outlines directions for future research.

\section{Related Work}\label{related_work}
Wayuunaiki is an Arawakan language primarily used within the Wayuu indigenous community and is spoken by approximately 420,000 people across northern Colombia and Venezuela. Additionally, in contrast to English, it features a predominant subject–object–verb (SOV) word order and exhibits agglutinative morphology, in which words are formed by combining morphemes, each contributing distinct semantic or grammatical information. However, despite its relatively large number of speakers compared to other indigenous languages in the region, Wayuunaiki remains underrepresented in the NLP field, with few applications and datasets available.

Most efforts to date have focused on developing linguistic resources—such as aligned sentence-pair corpora and descriptive analyses—and on building Wayuunaiki–Spanish translation systems. Notable examples include Rafael José Negrette Amaya’s bilingual Wayuunaiki–Spanish dictionary, which contains over 74,000 entries \cite{diccionario_wayuu}, and the aligned translations of religious and institutional texts, ranging from the Bible and the Colombian Constitution to various educational materials and linguistic studies of Wayuunaiki \cite{prieto-etal-2024-translation}. In terms of translation systems, key developments include the first Wayuunaiki–Spanish neural machine translation system built in 2023 \cite{graichen-etal-2023-enriching}; the fine-tuning of large Finnish-language pretrained models selected for their structural parallels to Wayuunaiki; and adaptations of multilingual frameworks such as Meta’s No Language Left Behind (NLLB) model, which supports numerous low-resource languages \cite{preserving_heritage2024, prieto-etal-2024-translation, hus-etal-2025-machine, nllbteam2022language}. 

While these efforts demonstrate that contemporary architectures can be adapted to Wayuunaiki–Spanish translation, published evaluations report modest performance, primarily due to the scarcity of parallel data and the narrow topical coverage of existing corpora \cite{graichen-etal-2023-enriching, hus-etal-2025-machine}. Moreover, training data frequently fail to reflect the language as it is actively spoken: in the AmericasNLP Shared Task, BLEU scores on up-to-date, carefully curated test sets differ markedly from those on standard validation sets, highlighting the need for novel, data-efficient modeling techniques and for resources that better capture real-world linguistic variation \cite{amercasNLP_2025}.

Recently, researchers have found that adopting RL techniques as an additional training stage for LLMs can significantly improve their performance, while requiring substantially less data than in the pre-training phase. Specifically, these advancements have been driven by two RL algorithms, PPO \cite{PPO2017}, which was used in the popular RLHF method \cite{RLFHF} to better align the output of models with user preferences; and GRPO \cite{grpo_deepseek2025, deepseek_r1, shao2024deepseekmathpushinglimitsmathematical}, introduced by DeepSeek to further enhance memory efficiency during RL-based training and to allow models to improve their coding, math, and reasoning capabilities.

In 2024, Zhan et. al \cite{zhang-etal-2024-reinforcement} introduced a reinforcement learning domain adaptation approach for neural machine translation, utilizing in-domain monolingual data to mitigate overfitting and reinforce domain-specific knowledge acquisition. Their method involves training a ranking-based model with a small-scale in-domain parallel corpus, which serves as a reward model to select higher-quality generated translations during fine-tuning.

Apart from the promise of RL techniques, agent-based frameworks have also been proposed to address the complexities of translation tasks. For instance, inspired by traditional human translation workflows, \citet{brivaiglesias2025aiagentsnewmachine} presented a multi-agent system for translating ultra-long literary texts, where specialized agents collaborate to handle different aspects of the translation process—such as adequacy review and fluency enhancement—resulting in translations that better maintain contextual fidelity and cultural nuances. While not specifically designed for translation tasks, other agent-based solutions have shown great potential by integrating external tools into LLMs, thus extending their abilities to perform more complex tasks. Recent approaches such as Search-R1 \cite{search-r1}, ReTool \cite{feng2025retool}, and SWiRL \cite{goldie2025syntheticdatageneration} even employ reinforcement learning to teach models when and how to use these external tools, which include code interpreters, calculators, or web search. However, despite being especially relevant for low-resource language translation tasks, where external tools like dictionaries could compensate for the limited training data, such agent-based methods remain underexplored in the translation domain.

\section{Methods}\label{methods}
\begin{figure*}[ht!]
    \centering
    \includegraphics[width=0.9\linewidth]{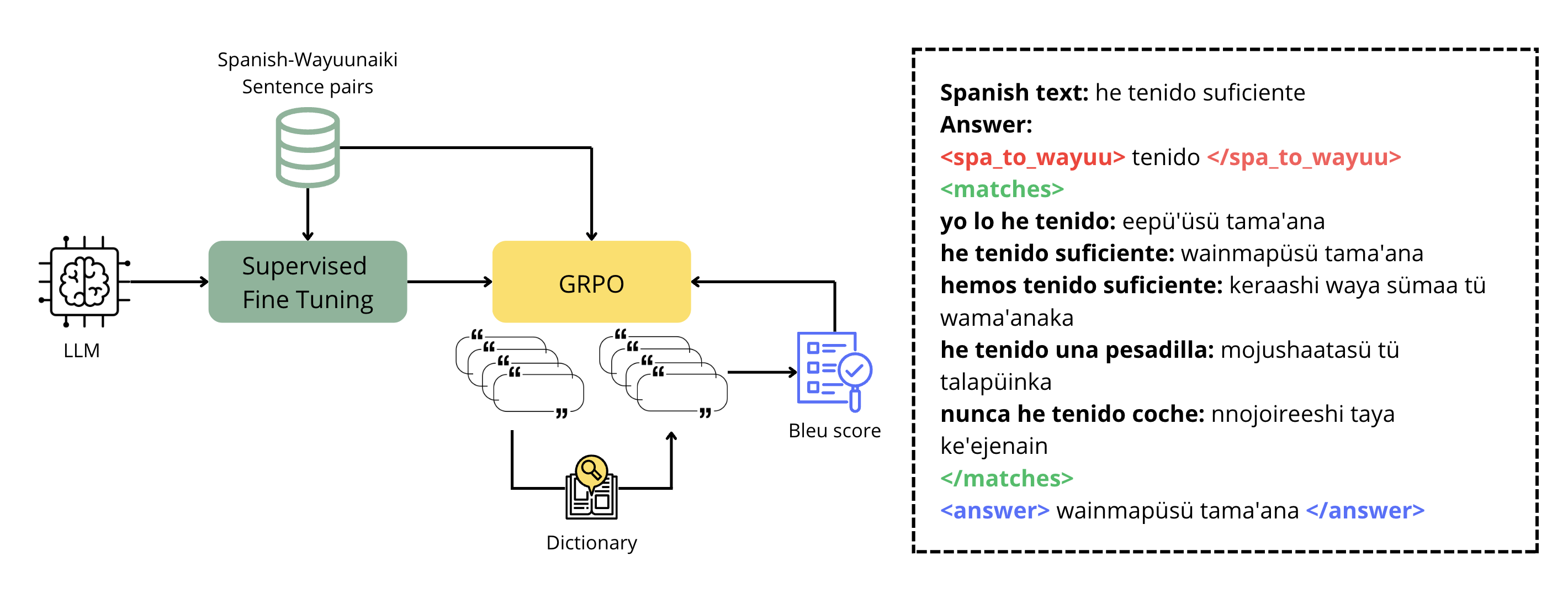}
     \caption{Overview of the training pipeline. A large language model  is first finetuned using supervised learning on Spanish–Wayuunaiki sentence pairs. The finetuned model is then further optimized using GRPO, where the reward is based on BLEU scores computed against reference translations. During this phase, the model can optionally use a dictionary tool to assist translation. The right-hand side illustrates an example of how the model interacts with the dictionary during the generation process.}
     \label{fig:methodology}
\end{figure*}

Figure \ref{fig:methodology} summarizes our methodology. To develop our translation system, we start  with an already pretrained large language model capable of following user instructions. After  selecting this base model, we perform supervised fine-tuning using an artificially augmented dataset that consist of Wayuunaiki-Spanish translation pairs and automatically generated examples of dictionary lookups. Finally, we use RL to boost the translation performance of our system.

\subsection{Supervised fine-tuning phase}

The supervised fine-tuning stage serves two key purposes: (1) to train the model to produce outputs in a structured format using predefined tags, and (2) to enable the model to learn how to properly invoke the dictionary tool. In this stage, we train the model on Spanish–Wayuunaiki translation examples using a prompt template that instructs the model how to invoke the dictionary tool and how to format its final translation (see Appendix A1).

As is common practice, the Spanish text and its corresponding Wayuunaiki translation are concatenated to the previous prompt to illustrate the translation task. To teach the model how to use the external dictionary tool, we insert artificial examples of dictionary calls immediately before the Wayuunaiki translation. To generate these examples, between zero and four words are randomly selected from the Spanish side to be queried using the dictionary tool. Then, for each lookup, the output of the dictionary—which consists of the first five matches from the dictionary entries—is also appended to the prompt.

Although these examples are randomly generated and are probably useless to achieve the correct translation, recent findings on the cognitive behaviors underlying self-improving reasoning in language models \cite{gandhi2025cognitive} suggest that acquiring structured habits, such as proper tool usage, can further enhance the performance achieved in the reinforcement learning stage. This benefit arises because the reinforcement learning phase can focus only on refining its tool usage rather than having to learn it entirely from scratch.

\subsection{Reinforcement learning phase}

Once the model has been fine-tuned to follow the structured prompt format and correctly use the dictionary tool, we proceed to the reinforcement learning stage. We adopt the GRPO framework \cite{deepseek_r1}, which is designed to align LLM behavior with complex tasks. In this setup, the language model itself acts as the policy. At each training step, we sample a Spanish–Wayuunaiki sentence pair and generate multiple candidate translations. Specifically, we generate 8 different translations for the same input prompt as defined during fine-tuning, which potentially include different combinations of dictionary tool invocations.

For each prediction, only the text enclosed within the \texttt{<answer>} tags is extracted and used for evaluation. Each generated output is then evaluated against a reference translation using BLEU \cite{bleu}, which serves as the reward signal for GRPO to update the policy based on translation quality. Additionally, tool outputs are masked to ensure they do not contribute to the policy loss \cite{search-r1}. This process enables the model to iteratively refine its translation strategy, improving overall performance while learning when and how to use the dictionary tool more effectively. To monitor progress during training, we evaluate the model every 50 steps on a fixed set of 640 sentence pairs sampled from the training dataset.

Since our task involves translating into Wayuunaiki, a language that differs significantly from the original training distribution of the model, we adopt the approach used in DAPO \cite{yu2025dapoopensourcellmreinforcement} and Dr.GRPO \cite{grpo_deepseek2025}, which relax the traditional GRPO constraint based on KL-divergence penalties. This adjustment is essential because the model must undergo substantial behavioral changes to produce coherent Wayuunaiki translations. Standard regularization methods that constrain the model to remain close to its initial policy would limit its ability to adapt effectively.

\subsection{Datasets and models}
For training, we use the Spanish–Wayuunaiki parallel corpus introduced by Prieto et. al \cite{prieto-etal-2024-translation}, which was included in the AmericasNLP 2025 Shared Task \cite{amercasNLP_2025}. This dataset was chosen because it provides a more natural and modern context for evaluation, rather than relying on translations of formal documents such as the Bible. 

To support tool-augmented translation, we incorporate a bilingual dictionary compiled by Rafael Jose Negrette Amaya \cite{diccionario_wayuu}, which originally contains approximately 74,000 Spanish–Wayuunaiki word and phrase pairs. To ensure tool responses remain concise and manageable, we filter this dictionary to retain only entries with five words or fewer on the Spanish side, resulting in a final dictionary of approximately 29,000 entries.

For testing, we employ a curated translation dataset consisting of the opening pages of Jules Verne’s \emph{Journey to the Center of the Earth
} \cite{verne1864journey}, translated into Wayuunaiki by the company Wayuunaiki Translation Services and funded by the Universidad de Los Andes. This dataset was used as the official test set in the AmericasNLP 2025 Shared Task, underscoring the importance of employing up-to-date, native-speaker translations, since training corpora (e.g., the Bible, the Colombian Constitution) often differ substantially from contemporary spoken usage. For more information on the datasets, see the Data Appendix. 

As a base instruction model, we use Qwen2.5-0.5B-Instruct \cite{qwen2025}, which offers multilingual support across more than 20 languages and is specifically optimized for cross-lingual tasks. One of the key design choices behind this model is its ability to generalize across languages through a cross-lingual transfer mechanism. This is achieved by translating instructions from high-resource languages into low-resource ones and generating corresponding response candidates. This training strategy makes Qwen2.5-0.5B-Instruct particularly well-suited for tasks involving low-resource languages such as Wayuunaiki, where robust generalization and instruction-following are essential.

\subsection{Training}\label{sec:training}

To evaluate model performance during training, we use the BLEU score \cite{bleu}, which measures translation quality by comparing overlapping n-grams between the generated output and a reference. For parameter-efficient adaptation, we apply LoRA (Low-Rank Adaptation) \cite{lora} in both supervised fine-tuning and reinforcement learning. In the RL phase, we further optimize for efficiency and stability by (1) leveraging vLLM \cite{vllm} for faster inference and trajectory sampling, (2) accumulating gradients over eight steps to balance memory footprint and effective batch size, (3) integrating DeepSpeed \cite{deepspeed} to reduce memory usage and boost throughput, and (4) omitting clipping in the policy loss, which allows us to keep only a single model instance in memory throughout training. All models are optimized with AdamW at a fixed learning rate of $5\times10^{-6}$.

\subsection{Experimental setup}

Our experiments systematically evaluate three key factors: training approach (zero-shot, supervised fine-tuning, reinforcement learning), dictionary access (available vs. unavailable), and model architecture (instruction-tuned vs. translation-specific models).

We begin by establishing baselines using the instruction-tuned model Qwen2.5-0.5B-Instruct in zero-shot settings. 

To test whether tool awareness alone is beneficial, we also include a variant where the model is informed that a dictionary is available but receives no examples of how to use it.

We then explore supervised fine-tuning to assess whether explicit demonstrations improve performance. One set of experiments uses standard parallel sentence pairs without tool interaction, serving to isolate the benefits of exposure to target-domain data. A second set extends this by introducing synthetic demonstrations that show the model how to use the dictionary tool. These examples are automatically constructed and illustrate when and how to query the tool during translation, allowing us to test whether models can learn tool-augmented behaviors from examples alone. For both settings, models were fine-tuned for one epoch on 59,715 paired sentences, using a learning rate of $1\times10^{-4}$, the AdamW optimizer, and prompt masking to ensure training focused only on the target completions.

We then evaluate a combined approach where SFT is followed by RL, in order to assess whether reinforcement learning can further refine tool usage and translation quality after initial supervised adaptation. These experiments are run both with and without tool access, allowing us to isolate the impact of the dictionary in the context of policy optimization. Notably, RL training for the tool-enabled model is performed on an SFT-trained version that incorporates tool usage, whereas for the tool-free model, RL is applied to an SFT-trained version that was not exposed to the tool.

Within the RL framework, we explore two reward strategies: sentence-level BLEU scores \cite{bleu} and character-level edit-based rewards \cite{morris2004CER}. Additionally, we examine the effect of RL training duration by directly comparing the performance of models trained for 400 steps versus those trained for 1400 steps.

Finally, to assess the generality of our approach, we replicate key experiments across different model architectures. We apply our full methodology—involving SFT and RL with dictionary access—to Llama-3.2-1B-Instruct, enabling a comparison over different pretraining bases. We also test a larger model, Qwen2.5-7B-Instruct, to explore whether scale offers measurable gains in low-resource translation. In parallel, we test our RL framework on a translation-specific model, NLLB \cite{nllbteam2022language}, which is not instruction-tuned and cannot utilize the tool. For this setup, we use the Wayuunaiki-specific checkpoint from \cite{prieto-etal-2024-translation} and apply GRPO without tool access or prompting, thereby isolating the effects of reinforcement learning on a model with strong translation priors.

To evaluate all our models, we use the average BLEU score computed between sentences on the 503 samples from the test set. Additionally, we measure different metrics to analyze tool usage. To ensure cost efficiency, we cap the number of allowed dictionary calls at a maximum of four.
\section{Results}\label{results}


This section presents the experimental results evaluating the performance of different models and training approaches for Spanish-to-Wayuunaiki translation, primarily using the BLEU score as the evaluation metric.

\begin{figure*}[htb!]
    \centering
    \includegraphics[width=0.8 \linewidth]{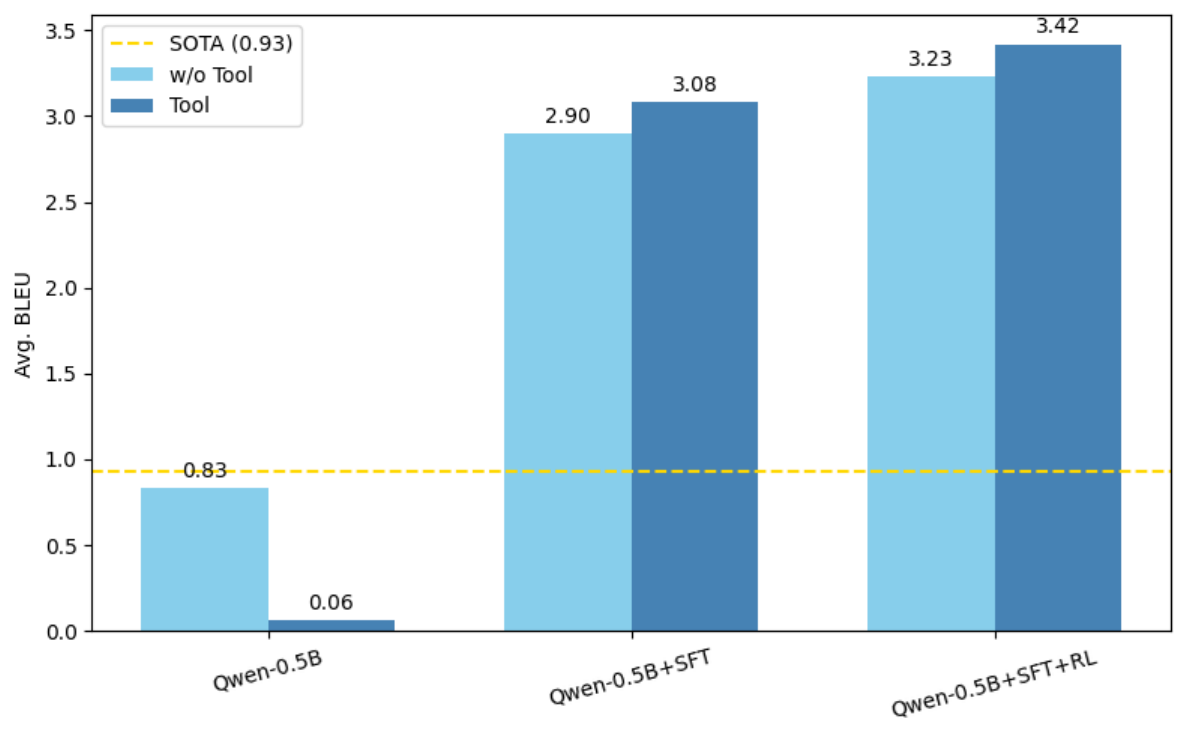}
    \caption{Average BLEU scores for different Qwen model variants, with and without tool usage. The results show that SFT effectively imparts basic translation capabilities, while RL yields a modest improvement on top of it. Enabling the dictionary tool provides an estimated 6\% relative gain.}
    \label{fig:bleu_chart}
\end{figure*}

Figure~\ref{fig:bleu_chart} presents the main results for the Qwen model under three configurations: without any fine-tuning (Base), with supervised fine-tuning (SFT), and with an additional reinforcement learning (RL) stage comprising 1,400 steps, using BLEU as the reward signal. The base Qwen-0.5B model achieved very low BLEU scores (0.83 without the tool, 0.06 with the tool)
, underscoring the need for training on Wayuunaiki data. Performance improved consistently at each stage of training, with SFT contributing the largest gain, and RL delivering an additional 11\% improvement, both with and without dictionary access. Additionally, the external dictionary tool provided a relative performance boost of approximately 6\% in both the SFT and SFT+RL stages. While prior work reported an average BLEU score of 10.54 on a test set similar to their training set \cite{preserving_heritage2024}, their model achieved only 0.93 BLEU on the curated test set used in our evaluation \cite{amercasNLP_2025}. These results demonstrate the effectiveness of our combined SFT and RL training pipeline, particularly when enhanced by access to an external dictionary tool.

Table~\ref{tab:tool_usage} offers a detailed breakdown of performance and tool usage across our training pipeline with the dictionary enabled. Notably, \textbf{the best-performing model (Qwen-0.5B+SFT+RL) makes the most extensive use of the dictionary}, employing it in every case and averaging 3.94 calls per sample, close to the allowed maximum of 4. The SFT stage plays a key role in enhancing performance by providing examples that teach the model both accurate translation pairs and effective tool usage. This is reflected in a success rate of almost 90\% when querying the dictionary, i.e., receiving valid matches for the queried word. These capabilities were further reinforced during the RL stage, which enabled the model to fully exploit the external tool, achieving a 95\% success rate.

\begin{table}[ht]
\centering
\small
\begin{tabular}{l@{\hspace{1pt}} c c c c}
\toprule
\textbf{Model} & \makecell{\textbf{Avg.}\\ 
\textbf{BLEU}} & \makecell{\textbf{Answers}\\ 
\textbf{w/ Tools}} & \makecell{\textbf{Avg.}\\ 
\textbf{Tool Calls}}  & \makecell{\textbf{Succ.}\\\textbf{Tool Calls}} \\
\midrule
Base & 0.06 & 45.72\% & 1.00 & 0.02\% \\
Base+SFT & 3.08 & 99.00\% & 2.13 & 89.76\% \\
Base+SFT+RL & \textbf{3.42} & \textbf{100.00\%} & \textbf{3.94} & \textbf{95.23\%} \\
\bottomrule\\
\end{tabular}
\caption{Tool usage and BLEU scores for different variants of the Qwen-0.5B model. The results indicate that better-performing models make more extensive use of the dictionary tool. Notably, the Qwen-0.5B+SFT+RL model invokes the tool in every response and approaches the maximum allowed number of calls per translation, averaging 3.94 out of 4.}
\label{tab:tool_usage}
\end{table}

Moreover, in Table~\ref{tab:bleu_tool_usage}, we evaluate our proposed method using different model architectures: Qwen2.5, LLaMA3.2, and NLLB. We also assess its effectiveness across different sizes of the Qwen model (0.5B and 7B parameters). For NLLB, which is not instruction-tuned, the dictionary tool is disabled. Additionally, the base NLLB model cannot be tested, as it does not natively support Wayuunaiki.

The results indicate that instruction-tuned models (Qwen and LLaMA) benefit significantly from both the SFT and SFT+RL stages when tool access is enabled. All instruction-tuned models achieve their best performance when trained using the complete pipeline. In contrast, the RL stage does not appear to enhance the performance of the NLLB model, which remains below that of the other tested models. Notably, with the exception of NLLB, larger models tend to achieve better results. Qwen2.5-7B reaches the highest average BLEU score of 4.45, outperforming all other models.

Tool usage also becomes more frequent and sophisticated across training stages, as models learn to more effectively leverage the dictionary. Since base larger models like Qwen2.5-7B are already capable of using the tool properly, tool usage does not necessarily increase in volume but becomes more refined, contributing to improved performance. A more detailed analysis of tool usage is provided in the following subsection.


\begin{table}[hbt!]
\centering
\small
\begin{tabular}{l c c c}
\toprule
\textbf{Model} & \textbf{Avg. BLEU} & \makecell{\textbf{Answers w/} \\ \textbf{Tools}} & \makecell{\textbf{Avg. Tool} \\ \textbf{Calls}} \\
\midrule

\multicolumn{4}{l}{\textbf{Base Models}} \\
Qwen-0.5B       & 0.06   & 45.72\%  & 1.00 \\
Llama3.2-1B     & 0.11  & 59.05\%  & 2.31 \\
Qwen-7B         & 2.10  & 94.04\%  & \textbf{4.22} \\
NLLB-3B            & --     & --       & --   \\

\midrule
\multicolumn{4}{l}{\textbf{+ SFT}} \\
Qwen-0.5B       & 3.08  & 99\%  & 2.13 \\
Llama3.2-1B     & 3.15  & 99\%  & 2.98 \\
Qwen-7B         & 4.33  & 97.81\%  & 2.97 \\
NLLB-3B            & 0.93      & --       & --   \\

\midrule
\multicolumn{4}{l}{\textbf{+ RL}} \\
Qwen-0.5B       & 3.16  & \textbf{100\%}  & 2.97 \\
Llama3.2-1B     & 3.48  & \textbf{100\%}  & 3.88 \\
Qwen-7B         & \textbf{4.45}  & 98.01\%  & 2.78 \\
NLLB-3B            & 0.93     & --       & --   \\

\bottomrule\\
\end{tabular}
\caption{Performance comparison across base models, SFT, and RL stages. Instruction-tuned models show significant improvements through both SFT and RL, partly due to their increasing use of external tools, as analyzed in the subsequent results subsection. Larger instruction-tuned models tend to perform better, with Qwen7B+SFT+RL achieving the highest score (4.45 Avg. BLEU), effectively doubling its base performance.}
\label{tab:bleu_tool_usage}
\end{table}

In Table~\ref{tab:rl_bleu}, we analyze the impact of different reward signals (BLEU versus CharacTer Error Rate) and the number of RL steps (400 vs. 1400) during the final RL training stage of the Qwen2.5-0.5B model. The results indicate that the BLEU metric is the only effective signal for improving the translation performanceof the model, yielding a 2.6\% improvement after 400 steps and achieving an 11\% relative gain with 1400 steps. In contrast, using the CharacTer metric leads to a 10.4\% performance degradation. Although there is some improvement after the initial 400 steps, the performance does not recover even after 1400 steps of training.

Despite the divergence in translation quality, both reward signals lead to increased tool usage over the course of RL training. The average number of tool calls per use rises from 2.13 to 3.94, and tool usage frequency increases from 99\% to 100\% after 1400 steps with both metrics.


\begin{table}[hbt!]
\centering
\small
\begin{tabular}{l c c c}
\toprule
\textbf{Reward Signal} & \textbf{Avg. BLEU} & \makecell{\textbf{Answers}\\ \textbf{w/ Tools}} & \makecell{\textbf{Avg.}\\ \textbf{Tool Calls}} \\
\midrule
Qwen-0.5+SFT & 3.08 & 99\% & 2.13 \\
\midrule
\multicolumn{4}{l}{\textbf{400 RL Steps}} \\
BLEU      & 3.16 & \textbf{100\%} & 2.97 \\
CharacTer & 2.59 & \textbf{100\%} & 3.02 \\
\midrule
\multicolumn{4}{l}{\textbf{1400 RL Steps}} \\
BLEU      & \textbf{3.42} & \textbf{100\%} & \textbf{3.94} \\
CharacTer & 2.76 & \textbf{100\%} & \textbf{3.94} \\
\bottomrule\\
\end{tabular}
\caption{Effect of reward signal type and RL training duration on BLEU scores and tool usage. The results show that BLEU scores outperform CharacTer scores as the reward signal. Increasing the number of RL training steps significantly improves performance and encourages more  intensive tool usage.}
\label{tab:rl_bleu}
\end{table}

\subsection{Dictionary Usage Analysis}

To evaluate how effectively the models leverage the dictionary tool, we measured the number of successful dictionary lookups for three versions: the base model, the model fine-tuned with SFT, and the model trained with both SFT and RL. As an upper bound, we defined a successful query as one where the Spanish word appears in the dictionary. Since the model is limited to querying a maximum of four words per sample, the theoretical maximum number of successful queries is 1,798.

\begin{figure}
    \centering
    \includegraphics[width=\linewidth]{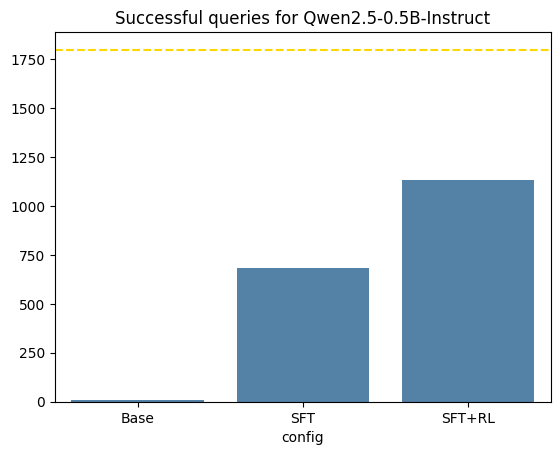}
    \caption{Number of dictionary lookups that returned results, referred to as successful queries. The results indicate that successful queries increase across training stages. The yellow horizontal line marks the theoretical upper bound of successful queries in our setup, which limited each sample to a maximum of 4 dictionary calls. Considering this constraint and filtering only for words present in the dictionary, the maximum achievable number of successful queries is 1798.}
    \label{dictionary_usage}
\end{figure}

As shown in Figure~\ref{dictionary_usage}, the number of successful lookups increases substantially at each training stage. The model trained with both SFT and RL achieved 1,130 successful lookups, a 65\% improvement over the model trained with SFT alone. These results highlight the effectiveness of each training phase in teaching the model to better utilize the dictionary tool to enhance translation performance. The fully trained model reaches 63\% of the theoretical maximum. However, it is important to note that, due to knowledge already acquired during the SFT phase, querying every word may not be necessary, as some words may already be known by the model.

Furthermore, we assessed the impact of dictionary integration on translation quality by comparing, for each lookup, the maximum BLEU score attainable using only the dictionary's best suggestion against the BLEU score of the final output of the model. For the SFT model, the mean “dictionary-only” BLEU is 0.109, whereas the mean BLEU of the model reaches 3.07; a paired two-sided t-test yields a $p$-value of \(p=6\times10^{-17}\), and 64\% of examples have a better BLEU score for the model output than for the best dictionary result. Similarly, the SFT+RL model attains a mean “dictionary-only” BLEU of 0.21 and a mean BLEU of 3.42 for the model's output ($p=4.6\times10^{-15}$), with improvements in 63.2\% of cases. These results demonstrate that the trained models, when using the dictionary, produce translations that are statistically significantly better than simply selecting the best dictionary result. This suggests that the models do not merely copy from the dictionary but effectively refine and enhance suggestions using their learned language knowledge.

Nevertheless, we identified important limitations in the dictionary itself. Only 10.4\% of the unique Spanish words in the test set appear as entries in the dictionary, and of these, just 16.3\% provide a Wayuunaiki translation that matches the reference. These limitations significantly reduce the potential benefit of integrating the dictionary, as it provides limited support to the model when processing the test samples.

\section{Discussion and future work}\label{discussion}

Our findings provide strong evidence that LLMs trained using SFT and RL to leverage external lexical resources, such as dictionaries, significantly improve translation performance in low-resource settings. These results were consistent across different model architectures, including LLaMA and Qwen, and across various model sizes.

Although our experiments focused exclusively on the Wayuunaiki language, the methodology is broadly applicable, as it does not rely on any language-specific techniques. As long as a dictionary is available, our approach can be readily extended to other languages. In fact, for non-agglutinative languages, the benefits could be even greater, since words in such languages are typically easier to translate independently. This contrasts with agglutinative languages like Wayuunaiki, where words are often formed by chaining multiple subwords, complicating the translation process.

Importantly, the improvements from our method are complementary to those achieved through traditional SFT on parallel corpora. This suggests a promising research direction for enhancing translation performance beyond the limitations imposed by the scarcity of parallel data.

Despite our success, we observed that the effectiveness of the dictionary tool was significantly constrained by both its limited coverage of the Wayuunaiki language and its overall quality. In many cases, the suggestions of the tool did not align with our reference translations. This underscores the critical need to develop high-quality, reliable external resources that can support language models in future work.

Our experiments also revealed that the effectiveness of the RL stage is highly dependent on the type of reward signal employed. This raises important questions about why the CharacTer reward signal (which focuses on character-level matches rather than word-level matches, like BLEU) was insufficient to drive improvements and, in some cases, even led to performance regressions. Future research could investigate the properties that make a reward function effective in the context of machine translation.

Another crucial consideration is the use of evaluation datasets with multiple reference translations. Such datasets can account for the various valid ways to express the same content, thereby enabling the design of more robust and representative reward signals.
\section{Data and software availability}
\label{sec:availability}
The algorithms and the datasets supporting the results presented in this article are available at \href{https://github.com/Manuel-2011/rl_translator}{RLTranslator}.
\section{Limitations}
\label{sec:limitations}
Our study presents a novel approach to low-resource machine translation for Spanish-to-Wayuunaiki, demonstrating state-of-the-art performance on the evaluated test set using a combination of Supervised Fine-Tuning (SFT) and Reinforcement Learning (RL) augmented with a dictionary tool. However, our experimental setup and analysis faced several significant limitations. All experiments were conducted on a \textbf{single server at Universidad de los Andes, equipped with 4 RTX6000 GPUs that were shared among numerous students} undertaking various Natural Language Processing experiments. This limited computational access, coupled with each Reinforcement Learning step \textbf{taking several minutes} due to the need for generating multiple rollouts and computing rewards, severely constrained the scale and duration of our training. While the training dataset contains approximately 59,715 paired sentences, the final RL configurations were trained for 1400 steps, and increasing steps further showed performance plateauing. This restriction meant we were \textbf{forced to train using only a portion of the available dataset}, as the limited number of RL steps prevented extensive exposure to the full data variability. Furthermore, a critical limitation affecting our analysis was the \textbf{inability to access a native Wayuunaiki speaking person}. While automatic metrics like BLEU were used for evaluation, these do not fully capture the nuances of translation quality, fluency, or cultural appropriateness for a language with distinct structures like Wayuunaiki. Therefore, a thorough \textbf{qualitative analysis of the generated translations by native speakers is still pending and remains highly desirable} for future work to better understand the practical utility and accuracy of our system for the Wayuu community and to support ongoing language revitalization efforts.

\bibliographystyle{plainnat} 
\bibliography{sample-base}

\newpage
\section{Appendix}

\subsection{A1. Prompt template for dictionary usage}

Below is the complete prompt used to instruct the model to translate a Spanish text into Wayuunaiki. The prompt also includes guidance on how to use the dictionary tool.

\texttt{
``Translate the following Spanish text into Wayuunaiki. Begin by identifying any words or phrases you're unsure how to translate. Then, you may look up those words using the dictionary tool by wrapping the Spanish word in <spa\_to\_wayuu> and </spa\_to\_wayuu>, and doing that for every unknown word. The dictionary will return matches enclosed in <matches> and </matches>. You can use the dictionary as many times as necessary.
Once you have all the information you need, provide the final translation enclosed in <answer> and </answer>. For example: <answer> xxx </answer>.
\\
Spanish text: \{\}''
}

\subsection{A2. Training hyperparameters}

Table~\ref{tab:hparams_sft} and Table~\ref{tab:hparams} list the hyperparameters used during the SFT and RL training stages, respectively. These values were not optimized but instead were selected based on commonly used settings reported in prior literature.

\begin{table*}[h]
\resizebox{\textwidth}{!} {
\begin{tabular}{lcc}
\toprule
\multicolumn{1}{c}{\textbf{Hyperparameter}} & \textbf{Definition}                                                                                    & \textbf{Value} \\ \midrule
max\_steps                                  & Maximum number of examples seen                                                                        & 1400          \\
sims\_per\_prompt                           & Simulations to calculate reward per example                                                            & 8              \\
policy\_lr                                  & Learning rate for the policy update                                                                    & 5e-6           \\
temperature                                 & Temperature of the LLM for generations                                                                 & 1.0            \\
max\_new\_tokens                            & Maximum tokens generated by the LLM                                                                    & 512            \\ \midrule
r                                           & Rank of the approximation matrices used for LoRA                                                       & 64             \\
lora\_alpha                                 & Scaling factor for LoRA approximation matrices                                                         & 64             \\
accum\_grad\_steps & Gradient accumulation steps & 8 \\

\midrule
optimizer                                   & type of optimizer                                                                                      & AdamW          \\
policy\_lr                                  & Learning rate of the optimizer                                                                         & 5e-6           \\
betas                                       & optimizer beta                                                                                         & (0.9, 0.999)   \\
eps                                         & optimizer eps                                                                                          & 1e-8           \\
weight\_decay                               & optimizer weight decay                                                                                 & 0.0            \\
gradient\_clipping                          & optimizer gradient clipping                                                                            & 0.1            \\ \bottomrule
\end{tabular}}
\caption{Hyperparameters used for RL training}
\label{tab:hparams}
\end{table*}

\begin{table*}[h]
\resizebox{\textwidth}{!} {
\begin{tabular}{lcc}
\toprule
\multicolumn{1}{c}{\textbf{Hyperparameter}} & \textbf{Definition}                                                                                    & \textbf{Value} \\ \midrule
num\_epochs & Epochs number & 1 \\
training\_samples                                  & Number of training samples                                                                        & 59,715          \\
batch\_size                                  & Batch size                                                                   & 16           \\
 \midrule
r                                           & Rank of the approximation matrices used for LoRA                                                       & 64             \\
lora\_alpha                                 & Scaling factor for LoRA approximation matrices                                                         & 64             \\

\midrule
optimizer                                   & type of optimizer                                                                                      & AdamW          \\
lr                                  & Learning rate                                                                         & 1e-4           \\
betas                                       & optimizer beta                                                                                         & (0.9, 0.999)   \\
eps                                         & optimizer eps                                                                                          & 1e-8           \\
weight\_decay                               & optimizer weight decay                                                                                 & 0.01            \\
 \bottomrule
\end{tabular}}
\caption{Hyperparameters used for SFT training}
\label{tab:hparams_sft}
\end{table*}

\subsection{A3. Computing Infrastructure}

Only one successful run was considered for each experiment. All experiments were conducted on a cluster equipped with four RTX 6000 GPUs, each with 48 GB of memory. The training process utilized the PyTorch and DeepSpeed libraries, while inference was performed efficiently using vLLM.

\subsection{Data Appendix}
The training dataset for this study was obtained from Prieto et al. \cite{prieto-etal-2024-translation}. The test dataset was used as the Wayuunaiki translation test set in the AmericasNLP 2025 Shared Task \cite{amercasNLP_2025} and is accessible via the \href{https://colombialanguages.virtual.uniandes.edu.co/}{Machine Learning for Indigenous Language Preservation project website}.

\end{document}